\documentclass[journal]{IEEEtran}
\usepackage{amsmath}
\usepackage{amssymb}
\usepackage{algorithm}
\usepackage{algorithmicx}
\usepackage{algpseudocode}
\usepackage{booktabs}
\usepackage{multirow}
\usepackage{color}
\usepackage{cite}
\usepackage{bm}
\usepackage{balance}
\usepackage{amssymb}
\usepackage{bbding}

\ifCLASSINFOpdf
\else
   \usepackage[dvips]{graphicx}
\fi
\usepackage{url}

\usepackage{graphicx}

\begin{document}

\title{Joint Multi-scale Gated Transformer and Prior-guided Convolutional Network for Learned Image Compression}

\author{Zhengxin Chen, Xiaohai He, \IEEEmembership{Member, IEEE}, Tingrong Zhang, Shuhua Xiong, and Chao Ren, \IEEEmembership{Member, IEEE}

\thanks{This work was supported in part by the National Natural Science Foundation of China under grants 62271336, 62211530110, and 62171304, in part by the Sichuan Province International Science and Technology Innovation Cooperation Fund under grant 2024YFHZ0289, and in part by the TCL Science and Technology Innovation Fund. \textit{(Corresponding author: Xiaohai He.)}}
\thanks{Z. Chen, X. He, T. Zhang, S. Xiong, and C. Ren are with the College of Electronics and Information Engineering, Sichuan University, Chengdu 610065, China (e-mail: zhengxinchen1994@gmail.com; hxh@scu.edu.cn; tr.zhang.scu@gmail.com; xiongshu@scu.edu.cn; chaoren@scu.edu.cn).}}

\markboth{}
{Shell \MakeLowercase{\textit{et al.}}: Bare Demo of IEEEtran.cls for IEEE Journals}
\maketitle

\begin{abstract}
Recently, learned image compression methods have made remarkable achievements, some of which have outperformed the traditional image codec VVC. The advantages of learned image compression methods over traditional image codecs can be largely attributed to their powerful nonlinear transform coding. Convolutional layers and shifted window transformer (Swin-T) blocks are the basic units of neural networks, and their representation capabilities play an important role in nonlinear transform coding. In this paper, to improve the ability of the vanilla convolution to extract local features, we propose a novel prior-guided convolution (PGConv), where asymmetric convolutions (AConvs) and difference convolutions (DConvs) are introduced to strengthen skeleton elements and extract high-frequency information, respectively. A re-parameterization strategy is also used to reduce the computational complexity of PGConv. Moreover, to improve the ability of the Swin-T block to extract non-local features, we propose a novel multi-scale gated transformer (MGT), where dilated window-based multi-head self-attention blocks with different dilation rates and depth-wise convolution layers with different kernel sizes are used to extract multi-scale features, and a gate mechanism is introduced to enhance non-linearity. Finally, we propose a novel joint Multi-scale Gated Transformer and Prior-guided Convolutional Network (MGTPCN) for learned image compression. Experimental results show that our MGTPCN surpasses state-of-the-art algorithms with a better trade-off between performance and complexity.
\end{abstract}

\begin{IEEEkeywords}
Deep learning, image compression, transformer, convolution.
\end{IEEEkeywords}

\IEEEpeerreviewmaketitle

\begin{figure*}[htbp]
\begin{center}
\includegraphics[width=\textwidth]{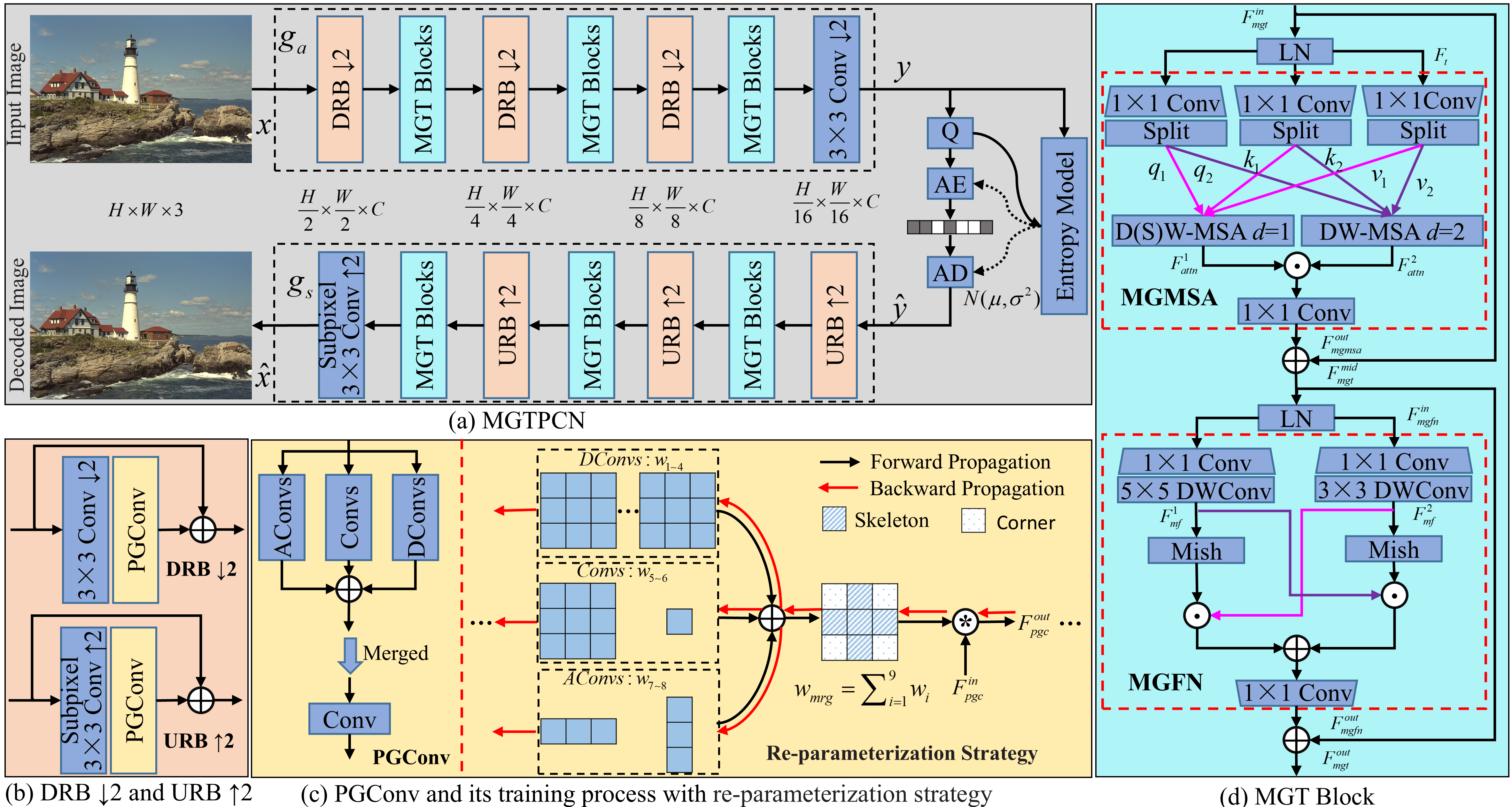}
\caption{The architecture of the proposed MGTPCN. $Q$ denotes quantization. AE and AD denote the arithmetic encoding and arithmetic decoding, respectively. }
\label{fig1}
\end{center}
\end{figure*}

\section{Introduction}
\IEEEPARstart{I}{mage} compression techniques are applied to various electronic devices and systems to achieve efficient data transmission and storage. Learned image compression methods \cite{1,2,3,4,5,6,7,9,10,11,12,13,15,30,31,32} have made remarkable achievements in recent years, some of which \cite{9,10,11,12,13,15} have outperformed the traditional image codec VVC \cite{16}. One of the important advantages of the learned image compression methods over the traditional image codecs is their powerful nonlinear transform coding \cite{17}. On the one hand, the powerful nonlinear transform can effectively remove the spatial redundancy of the input image and map it into a compact latent representation, reducing the bits required to encode the image. On the other hand, the powerful nonlinear transform ensures that accurate texture is recovered from the quantized latent representation to reconstruct a high-fidelity decoded image. Therefore, the powerful nonlinear transform plays an important role in improving the rate-distortion (RD) performance of learned image compression methods.

To improve the RD performance of learned image compression methods, researchers have proposed many nonlinear transforms based on neural networks \cite{1,2,3,5,6,7,9,10,11,12,13,15,30,31,32}, where Convolutional layers and shifted window transformer (Swin-T) blocks \cite{18} are widely used to extract local and non-local features. In early works \cite{1,2,3,5}, convolutional layers were used for nonlinear transform, and good RD performance was achieved. In recent years, to capture long-range dependencies, Zhu et al. \cite{7}, Zou et al. \cite{11}, and Liu et al. \cite{12} introduced the Swin-T block into the nonlinear transform network and further boosted the RD performance of the learned image compression methods. However, the vanilla convolution and Swin-T have some limitations, which may restrict the RD performance. In a $3\times 3$ window, vanilla convolutions tend to treat elements and weights at different positions equally, which may cause the following issues: 1) Important elements (e.g., skeleton elements) are ignored, while minor elements (e.g., corner elements) are over-emphasized. 2) The low-frequency information is over-emphasized, while the high-frequency information is ignored. For the Swin-T block, which contains a (shifted)window-based multi-head self-attention ((S)W-MSA) and a feed-forward network (FFN), the single-range receptive field restricts its ability to capture multi-scale contexts. Moreover, (S)W-MSA is essentially a linear operation, which performs a weighted sum on values based on the attention map calculated from queries and keys. The inherent linearity of the (S)W-MSA may make it hard to handle complex patterns.

To solve these issues, we propose a novel prior-guided convolution (PGConv), which contains parallel vanilla convolutions, asymmetric convolutions (AConvs), and difference convolutions (DConvs) \cite{19,20}. AConvs and DConvs are introduced to strengthen skeleton elements and high-frequency components, respectively. To reduce the computational complexity, a re-parameterization strategy is used to integrate parallel convolutions into one. Moreover, we propose a novel multi-scale gated transformer (MGT), where dilated (shifted)window-based multi-head self-attention (D(S)W-MSA) blocks with different dilation rates and depth-wise convolution (DWConv) layers with different kernel sizes are used to extract multi-scale features, and a gate mechanism is introduced to enhance non-linearity.

Overall, our contributions are three-fold:

1)	A novel PGConv combined with a re-parameterization strategy is proposed to improve the ability of the vanilla convolution to extract local features without increasing the computational complexity.

2)	A novel MGT is proposed to extract non-local features with multi-scale receptive fields and enhanced non-linearity.

3)	We develop a novel joint Multi-scale Gated Transformer and Prior-guided Convolutional Network (MGTPCN) for learned image compression, which surpasses state-of-the-art algorithms with a better trade-off between performance and complexity.

\section{Proposed Method}
\subsection{Overall Architecture}
Fig. \ref{fig1}(a) illustrates the overall architecture of the proposed MGTPCN, which contains three key parts: an analysis transform ${{g}_{a}}$, a synthesis transform ${{g}_{s}}$, and an entropy model. The analysis transform is used to map the input image $x$ to the latent representation $y$, which is then sent to the quantizer $Q$ to obtain the quantized latent representation $\hat{y}$. The synthesis transform is used to inversely map the quantized latent representation $\hat{y}$ to the decoded image $\hat{x}$. In ${{g}_{a}}$, the spatial resolution of the input image is reduced by DRB$\downarrow$2 and $3\times 3$ Conv$\downarrow$2 to eliminate redundancy. In ${{g}_{s}}$, the spatial resolution of the quantized latent representation is increased by URB$\uparrow$2 and Subpixel $3\times 3$ Conv$\uparrow$2 to restore size. The symbols $\downarrow$ and $\uparrow$ represent downsampling and upsampling, respectively. As shown in Fig. \ref{fig1}(b), each DRB$\downarrow$2 comprises a $3\times 3$ Conv$\downarrow$2 and a PGConv, and each URB$\uparrow$2 comprises a Subpixel $3\times 3$ Conv$\uparrow$2 and a PGConv. Each Subpixel $3\times 3$ Conv$\uparrow$2 comprises a $3\times 3$ Conv and a Subpixel layer \cite{21}. MGT and PGConv will be introduced in detail in the following subsections. The main contribution of this paper is to propose a novel nonlinear transform (i.e., ${{g}_{a}}$ and ${{g}_{s}}$), leading to a compact latent representation $y$ and a high-fidelity reconstruction result $\hat{x}$. Therefore, following \cite{30}, the entropy model that combines spatial-channel context \cite{30} and hyper-prior \cite{2} modules is used to model $p(\hat{y})$ as Gaussian distribution. Our MGTPCN is trained using the loss function:
\begin{equation}
\label{eq1}
L=R(\hat{y})+R(\hat{z})+\lambda D(x,\ \hat{x}),
\end{equation}
where $R(\hat{y})=E[-{{\log }_{2}}({{p}_{\hat{y}|\hat{z}}}(\hat{y}|\hat{z})]$, $R(\hat{z})=E[-{{\log }_{2}}({{p}_{{\hat{z}}}}(\hat{z}))]$, the side information $\hat{z}$ is extracted by the hyper-prior module to help model the distribution $p(\hat{y})$, and the Lagrange multiplier $\lambda $ is used to balance the rate $R(\cdot )$ and distortion $D(\cdot )$.

\subsection{Prior-guided Convolution}
In a $3\times 3$ window, the central element usually enjoys a higher correlation with its horizontal and vertical elements (i.e., skeleton elements), which are more important to the central pixel and deserve more attention. In addition, like other low-level visual tasks (e.g., image dehazing \cite{20} and image super-resolution \cite{33}), high-frequency information plays an important role in image compression. In the analysis transform, although the downsampling operation can effectively reduce the spatial redundancy, it may remove useful high-frequency information, which is important for the synthesis transform to reconstruct a high-fidelity decoded image. These image priors inspire us to propose a novel PGConv. Fig. \ref{fig1}(c) shows the structure of PGConv, which consists of three types of parallel convolutions: Convs, AConvs, and DConvs. They can be merged into a common convolution by a re-parameterization strategy, which will be introduced in the following. Convs comprise two convolutions of kernel sizes $3\times 3$ and $1\times 1$. AConvs comprise two convolutions of kernel sizes $3\times 1$ and $1\times 3$, which guide PGConv to enhance the horizontal and vertical elements of a $3\times 3$ window, respectively. DConvs comprise central difference convolution \cite{19}, angular difference convolution \cite{19}, horizontal difference convolution \cite{20}, and vertical difference convolution \cite{20}, which guide PGConv to extract high-frequency information. The image priors introduced by AConvs and DConvs impose constraints on the solution space, guiding the optimization of convolutional weights towards stronger model representation capabilities. Given the input feature $F_{pgc}^{in}$, the output feature of PGConv is formulated as
\begin{equation}
\label{eq6}
F_{pgc}^{out}=\sum\nolimits_{i=1}^{8}{F_{pgc}^{in}*{{w}_{i}}},
\end{equation}
where $*$ denotes the convolution operation, and ${{w}_{i=1\sim 8}}$ denote the weights of two Convs, two AConvs, and four DConvs. Deploying multiple parallel convolutions inevitably increases the computational complexity. Fortunately, according to the additivity of the convolution operation \cite{20,24}, Eq. \ref{eq6} can be rewritten as
\begin{equation}
\label{eq7}
F_{pgc}^{out}={F_{pgc}^{in}}*\sum\nolimits_{i=1}^{8}{{{w}_{i}}}=F_{pgc}^{in}*{{w}_{mrg}},
\end{equation}
where ${{w}_{mrg}}$ denotes the merged weights, which combine multiple parallel convolutions into one. According to Eq. \ref{eq7}, we adopt a re-parameterization strategy to accelerate the training of PGConv. As shown in Fig. \ref{fig1}(c), during forward propagation, the weights of the parallel convolutions are summed at the corresponding positions to obtain the merged weights, which are used to convolve the input features. During backward propagation, the weights of the parallel convolutions are updated individually using the chain rule and gradient descent algorithms. Once the training is complete, the merged weights are used for inference and achieve the same results as parallel convolutions. Therefore, the re-parameterization strategy allows PGConv to achieve better performance than Conv without increasing the model complexity.

\subsection{Multi-scale Gated Transformer}

As shown in Fig. \ref{fig1}(d), our MGT block can be expressed as
\begin{equation}
\label{eq2}
\begin{aligned}
  & F_{mgt}^{mid}={{g}_{mgmsa}}({{g}_{\ln }}(F_{mgt}^{in}))+F_{mgt}^{in}, \\
 & F_{mgt}^{out}={{g}_{mgfn}}({{g}_{\ln }}(F_{mgt}^{mid}))+F_{mgt}^{mid}, \\
\end{aligned}
\end{equation}
where $F_{mgt}^{in}$,$F_{mgt}^{mid}$, and $F_{mgt}^{out}$ are the input, intermediate, and output features of the MGT block, respectively, and ${{g}_{\ln }}$, ${{g}_{mgmsa}}$, and ${{g}_{mgfn}}$ denote the functions of layer normalization (LN), MGMSA, and MGFN, respectively. The following elaborates on MGMSA and MGFN.

\begin{figure}
\centering
\includegraphics[width=\columnwidth]{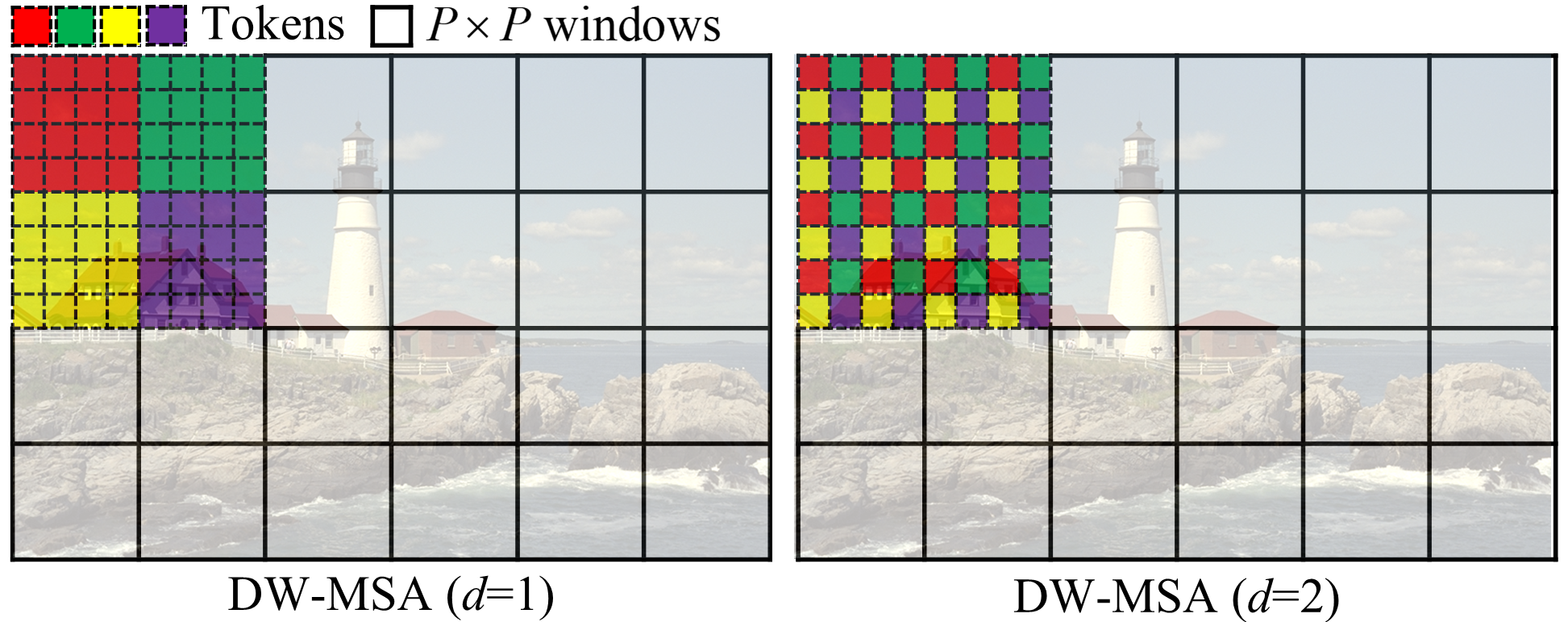}
\centering
\caption{DW-MSAs with different dilation rates.}
\label{fig2}
\end{figure}

\textit{1) MGMSA:} Two DW-MSAs with different dilation rates are introduced for efficient multi-scale feature extraction. As shown in Fig. \ref{fig2}, the DW-MSA divides the input features into non-overlapping windows of size $P\times P$. Similar to dilated convolution \cite{22}, DW-MSA inserts intervals between the tokens in each window to enlarge its receptive field and performs self-attention between the tokens with the same color. The dilation rate $d$ is introduced to control the interval size between tokens. It should be noted that when $d = 1$, D(S)W-MSA degenerates to vanilla (S)W-MSA. For the tokens ${{F}_{t}}\in {{R}^{{{P}^{2}}\times C}}$ with a certain color, their queries, keys, and values are first calculated by $1\times 1$ convolutions and then are split into two parts along the channel dimension:
\begin{equation}
\label{eq3}
{{q}_{1}},\ {{q}_{2}},{{k}_{1}},\ {{k}_{2}},{{v}_{1}},\ {{v}_{2}}={{g}_{s}}(g_{conv}^{1\times 1}({{F}_{t}})),
\end{equation}
where $g_{conv}^{1\times 1}$ and ${{g}_{s}}$ denote $1\times 1$ convolution and channel splitting operations, respectively. Subsequently, the split queries, keys, and values are fed into two DW-MSAs with different dilation rates, and the obtained multi-scale attention features $F_{attn}^{1}$ and $F_{attn}^{2}$ are element-wise multiplied with each other to perform the gate mechanism. Finally, a $1\times 1$ convolution further refines the gated feature to obtain the output $F_{mgmsa}^{out}$.
\begin{equation}
\label{eq4}
\begin{aligned}
  & F_{mgmsa}^{out}=g_{conv}^{1\times 1}(F_{attn}^{1}\odot F_{attn}^{2}), \\
 & F_{attn}^{i}=soft\max ({{{q}_{i}}k_{i}^{\top }}/{\tau }\;+B){{v}_{i}},\ i\in \{1,2\}, \\
\end{aligned}
\end{equation}
where $\odot $, $\tau $, and $B$ denote the element-wise multiplication operation, learnable scaling factor, and relative position encoding, respectively. The gate mechanism imposes nonlinearity and filters the less informative features, which helps the model learn complex patterns. However, from another perspective, this may hamper the flow of information in the network. Besides, the channel splitting operation halves the feature dimension. To maintain the feature dimension and promote the flowability of the feature information, the number of output channels of the $1\times 1$ convolution for queries is set to $2C$. Since queries and keys usually have dimensional redundancy \cite{23}, the number of output channels of the $1\times 1$ convolutions for them is set to half of their input embedding dimension, which reduces the model complexity.

\textit{2) MGFN:} Our MGFN adopts a dual parallel path, in which $1\times 1$ convolutions are first used to double the dimension of the input feature $F_{mgfn}^{in}$, and then DWConvs of kernel sizes $3\times 3$ and $5\times 5$ are used to extract the multi-scale features ${F_{mf}^{1}}$ and ${F_{mf}^{2}}$. Subsequently, the gate mechanism is performed by the element-wise multiplication between ${F_{mf}^{1}}$ and ${F_{mf}^{2}}$. Finally, an element-wise sum operation fuses the gated multi-scale features of different paths, and a $1\times 1$ convolution halves the dimension of the fused features to obtain the output $F_{mgfn}^{out}$.

\begin{equation}
\label{eq5}
\begin{aligned}
  & F_{mgfn}^{out}=g_{conv}^{1\times 1}(\sigma ({F_{mf}^{1}})\odot {F_{mf}^{2}}+\sigma ({F_{mf}^{2}})\odot {F_{mf}^{1}}), \\
 & {F_{mf}^{i}}=g_{dwc}^{(2i+1)\times (2i+1)}(g_{conv}^{1\times 1}(F_{mgfn}^{\operatorname{int}})),\ i\in \{1,2\}, \\
\end{aligned}
\end{equation}
where $\sigma $ and $g_{dwc}^{(2i+1)\times (2i+1)}$ denote Mish activation function and DWConv of kernel size $(2i+1)\times (2i+1)$, respectively.

\section{Experimental Results}
\subsection{Implementation Details}
The training dataset consists of $3\times {{10}^{5}}$ images from the OpenImages database \cite{25}. During the training phase, 16 random patches of resolution $256\times 256$ are cropped from these images. We utilize Kodak \cite{26} (24 images of size $768\times 512$) and Tecnick \cite{27} (100 images of size $1200\times 1200$) as the testing datasets. Peak signal-to-noise ratio (PSNR) and bits per pixel (bpp) are used to evaluate compression distortion and bit rate, respectively. The Bjontegaard delta rate (BD-Rate) \cite{28} is used to evaluate the RD performance of different algorithms. Besides, we employ the number of parameters and multiply-add operations (computed on a $768\times 512$ image) to compare their computational complexity.

We set the hyper-parameters $C = 192$ and $P = 8$. $D(\cdot )$ is implemented as the mean square error function. Following CompressAI \cite{29}, $\lambda $ is chosen from a predefined set of values: $\{18,25,35,67,130,250,483\}\times {{10}^{-4}}$. The model parameters are updated by the Adam algorithm with an initial learning rate of ${{10}^{-4}}$, which is halved if no performance improvement is observed over five consecutive epochs.

\begin{table}[htbp]
  \setlength\tabcolsep{4.5pt}
  \centering
  \caption{BD-rate results of different variations of the proposed MGTPCN on the Kodak dataset.}
    \begin{tabular}{cccccccc}
    \toprule
    \multicolumn{2}{c}{Modules} & ${{V}_{1}}$ & ${{V}_{2}}$ & ${{V}_{3}}$ & ${{V}_{4}}$ & ${{V}_{5}}$ & MGTPCN \\
    \midrule
    \multirow{3}[6]{*}{PGConv} & Convs & \checkmark     & \checkmark     & \checkmark     & \checkmark     & \checkmark     & \checkmark \\
\cmidrule{2-8}          & DConvs & $\times$      & \checkmark     & \checkmark     & \checkmark     & \checkmark     & \checkmark \\
\cmidrule{2-8}          & AConvs & $\times$      & $\times$      & \checkmark     & \checkmark     & \checkmark     & \checkmark \\
    \midrule
    \multirow{2}[4]{*}{Swin-T} & (S)W-MSA & $\times$      & $\times$      & $\times$      & \checkmark     & $\times$      & $\times$ \\
\cmidrule{2-8}          & FFN   & $\times$      & $\times$      & $\times$      & \checkmark     & $\times$      & $\times$ \\
    \midrule
    \multirow{2}[4]{*}{GT} & GMSA  & $\times$      & $\times$      & $\times$      & $\times$      & \checkmark     & $\times$ \\
\cmidrule{2-8}          & GFFN  & $\times$      & $\times$      & $\times$      & $\times$      & \checkmark     & $\times$ \\
    \midrule
    \multirow{2}[2]{*}{MGT} & MGMSA & $\times$      & $\times$      & $\times$      & $\times$      & $\times$      & \checkmark \\
\cmidrule{2-8}          & MGFN  & $\times$      & $\times$      & $\times$      & $\times$      & $\times$      & \checkmark \\
    \midrule
    \multicolumn{2}{c}{BD-Rate (\%)} & 0.00  & -1.73  & -3.39  & -7.77  & -9.21  & -11.64  \\
    \bottomrule
    \end{tabular}%
  \label{tab1}%
\end{table}%

\begin{figure}
\centering
\includegraphics[width=\columnwidth]{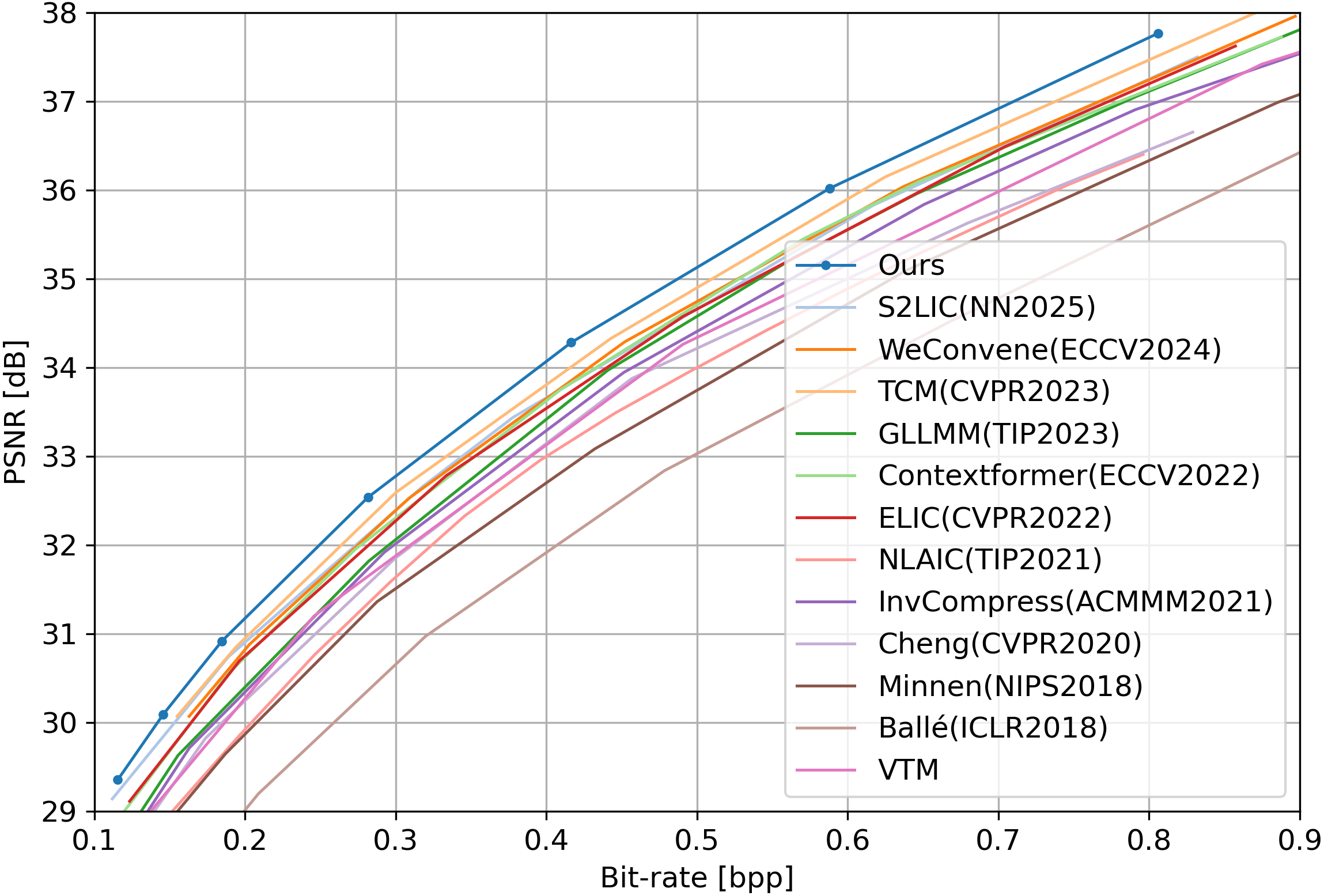}
\centering
\caption{RD curves of different methods on the Kodak dataset.}
\label{fig3}
\end{figure}

\begin{figure}
\centering
\includegraphics[width=\columnwidth]{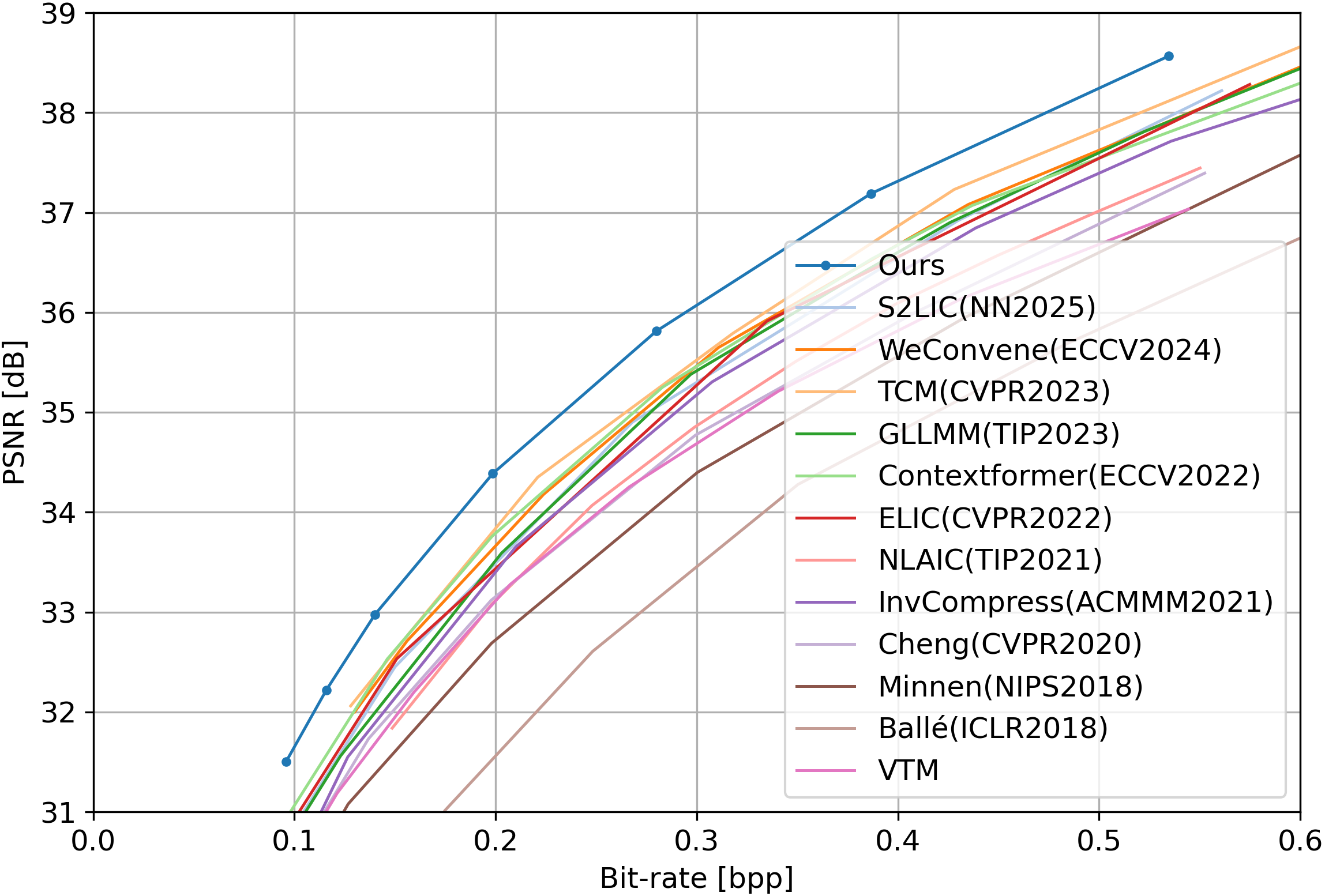}
\centering
\caption{RD curves of different methods on the Tecnick dataset.}
\label{fig4}
\end{figure}
\subsection{Model Analysis}
We construct five variations of MGTPCN to validate the effectiveness of PGConv and MGT. By replacing DW-MSA with W-MSA and removing DWConvs, MGMSA and MGFN degenerates to GMSA and GFFN, respectively, MGT degenerates to GT by replacing MGMSA and MGFN with GMSA and GFFN, and MGTPCN degenerates to ${{V}_{5}}$ by replacing MGT with GT. By removing the gate mechanism, GT, GMSA, and GFFN degenerate into Swin-T, (S)WMSA, and FFN, respectively, and ${{V}_{5}}$ degenerates into ${{V}_{4}}$ by replacing GT with Swin-T. ${{V}_{4}}$ degenerates to ${{V}_{3}}$ by replacing Swin-T with vanilla convolution. AConvs are removed from ${{V}_{3}}$ to generate ${{V}_{2}}$, and ${{V}_{2}}$ degenerates to ${{V}_{1}}$ by removing DConvs. Table \ref{tab1} reports the BD-rate results of these variations on the Kodak dataset. ${{V}_{1}}$ is set as the baseline. The BD-rate reductions of ${{V}_{2}}$, ${{V}_{3}}$, ${{V}_{4}}$, ${{V}_{5}}$, and MGTPCN over ${{V}_{1}}$ are 1.73$\%$, 3.39$\%$, 7.77$\%$, 9.21$\%$, and 11.64$\%$, respectively. The increasing BD-rate reduction demonstrates that DCnvs, AConvs, Swin-T, gate mechanism, and multi-scale feature extraction can all contribute to the RD performance of our MGTPCN. In other words, thanks to the proposed PGConv and MGT, the nonlinear transform of our MGTPCN enjoys powerful capabilities to remove the feature redundancy and reconstruct high-fidelity decoded images.

\begin{table}[htbp]
  \setlength\tabcolsep{1.2pt}
  \centering
  \caption{BD-rate results of different methods over VTM-18.0 Intra on the Kodak and Tecnick datasets.}
    \begin{tabular}{ccccc}
    \toprule
    \multirow{2}[4]{*}{Methods} & \multirow{2}[4]{*}{Params(M)} & \multirow{2}[4]{*}{Mult-Adds(G)} & \multicolumn{2}{c}{BD-Rate(\%)} \\
\cmidrule{4-5}          &       &       & Kodak & Tecnick \\
    \midrule
    VTM-18.0 Intra \cite{16} & -     & -     & 0.00  & 0.00  \\
    \midrule
    Ball\'{e}(ICLR2018) \cite{2} & 11.58  & 164.34  & 28.77  & 43.55  \\
    \midrule
    Minnen(NIPS2018) \cite{3} & 20.15  & 176.79  & 12.16  & 15.57  \\
    \midrule
    Cheng(CVPR2020) \cite{5} & 27.55  & 403.27  & 4.54  & 6.43  \\
    \midrule
    NLAIC(TIP2021) \cite{6} & -  & -  & 9.44  & 5.62  \\
    \midrule
    InvCompress(ACMMM2021) \cite{9} & 47.55  & 407.30  & -0.83  & -1.66  \\
    \midrule
    Contextformer(ECCV2022) \cite{31} & -  & -  & -7.12  & -8.65  \\
    \midrule
    ELIC(CVPR2022) \cite{30} & 31.66  & 326.78  & -7.08  & -8.01  \\
    \midrule
    GLLMM(TIP2023) \cite{10} & -  & -  & -4.18  & -7.22  \\
    \midrule
    TCM(CVPR2023) \cite{12} & 75.90  & 700.66  & -11.88  & -12.97  \\
    \midrule
    WeConvene(ECCV2024) \cite{13} & 105.51  & 904.56  & -8.51  & -8.88  \\
    \midrule
    S2LIC(NN2025) \cite{32} & -  & -     & -9.93  & -7.23  \\
    \midrule
    \textbf{Ours} & \textbf{66.32 } & \textbf{457.76 } & \textbf{-16.67 } & \textbf{-22.18 } \\
    \bottomrule
    \end{tabular}%
  \label{tab2}%
\end{table}%

\begin{figure*}
\centering
\includegraphics[width=\textwidth]{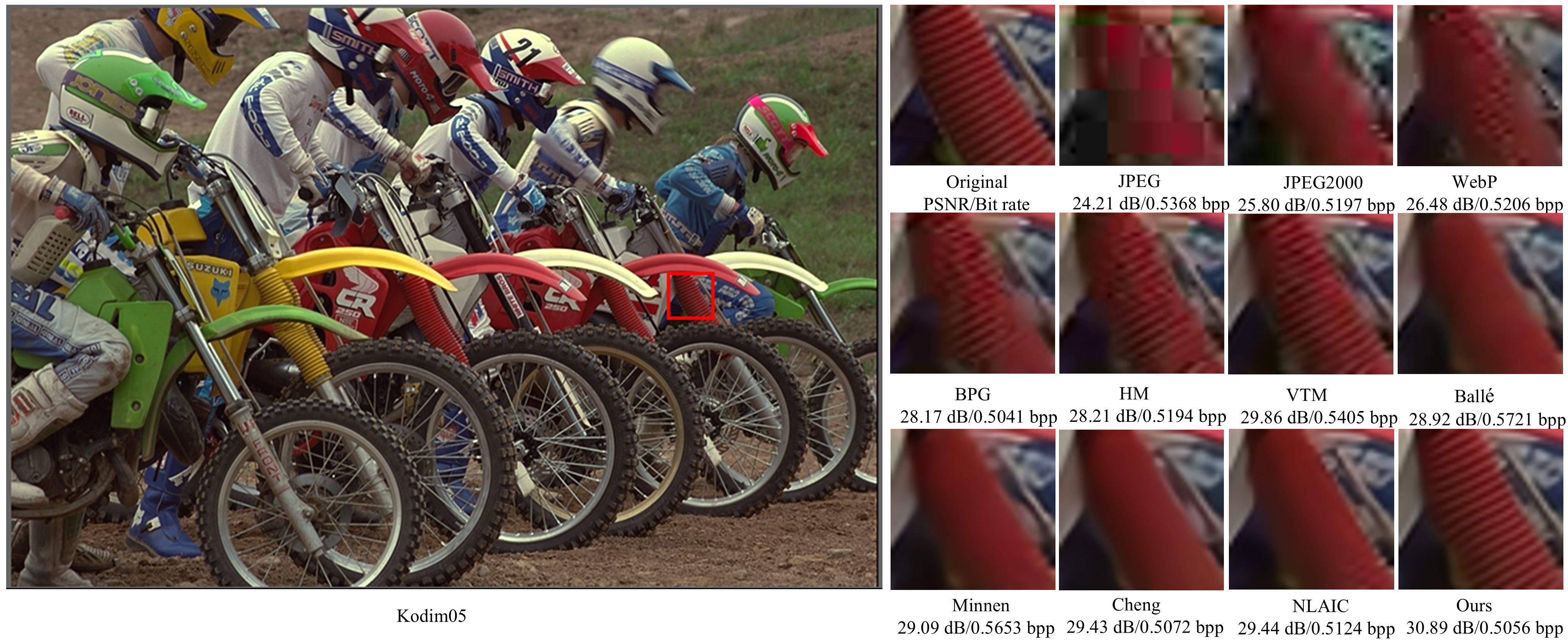}
\centering
\caption{Visualization of the decoded image from the Kodak dataset.}
\label{fig5}
\end{figure*}
\subsection{Rate-Distortion Performance}

Our MGTPCN is compared with several state-of-the-art methods, including the learning-based methods Ball\'{e} \cite{2}, Minnen \cite{3}, Cheng \cite{5}, NLAIC \cite{6}, InvCompress \cite{9}, ELIC \cite{30}, Contextformer \cite{31}, GLLMM \cite{10}, TCM \cite{12}, WeConvene \cite{13}, S2LIC \cite{32}, and the traditional codecs BPG \cite{34} and VVC (VTM-18.0 Intra). Their RD curves on the Kodak and Tecnick datasets are illustrated in Fig. \ref{fig3} and Fig. \ref{fig4}, respectively. As we can see, our MGTPCN outperforms competing methods, delivering higher PSNR values at the same bit-rate. Quantitative comparisons are provided in Table \ref{tab2}, where the BD-rate results are computed with VTM serving as the baseline. The number of parameters and multiply-add operations of different methods are also reported. Notably, our MGTPCN demonstrates superior RD performance, achieving significant BD-rate reductions. For instance, on the Tecnick dataset, the BD-rate saving produced by our MGTPCN is 22.18$\%$, while that of TCM is only 12.97$\%$. Overall, our MGTPCN achieves a better balance between RD performance and model complexity in terms of both the BD-rate and the number of parameters and multiply-add operations.

Fig. 5 visualizes the decoded images of different methods. As we can see, our MGTPCN yields the decoded image with higher fidelity and lower bit-rate.

\section{Conclusion}
In this paper, we propose a novel MGTPCN for learned image compression, where PGConv and MGT are developed to improve nonlinear transform coding. Compared with vanilla convolution, our PGConv combined with a re-parameterization strategy enjoys a stronger ability to extract local features without increasing the model complexity. Compared with Swin-T, our MGT enjoys a stronger ability to extract non-local features with multi-scale receptive fields and enhanced non-linearity. Experimental results show the effectiveness of our PGConv and MGT and the superiority of our MGTPCN over state-of-the-art methods. How to combine PGConv and MGT in an efficient manner is our future work.

\balance

\bibliography{mybibdatabase}
\bibliographystyle{IEEEtran}

\end{document}